\documentclass[10pt,twocolumn,letterpaper]{article}

\usepackage{wacv}
\usepackage{times}
\usepackage{epsfig}
\usepackage{graphicx}
\usepackage{amsmath}
\usepackage{amssymb}
\usepackage{booktabs}
\usepackage{authblk}
\usepackage{color}
\usepackage{multirow}
\usepackage{enumitem}
\usepackage[labelformat=simple]{subcaption}

\usepackage{array}
\newcommand{\PreserveBackslash}[1]{\let\temp=\\#1\let\\=\temp}
\newcolumntype{C}[1]{>{\PreserveBackslash\centering}p{#1}}
\usepackage{makecell}

\def\wacvPaperID{1716} 

\wacvapplicationstrack 

\wacvfinalcopy 

\ifwacvfinal
\usepackage[breaklinks=true,bookmarks=false]{hyperref}
\else
\usepackage[pagebackref=true,breaklinks=true,colorlinks,bookmarks=false]{hyperref}
\fi

\begin{document}

\title{Sim2real Transfer Learning for Point Cloud Segmentation: An Industrial Application Case on Autonomous Disassembly \vspace{-0.2cm}}

\author[1]{Chengzhi Wu}
\author[1]{Xuelei Bi}
\author[2,3]{Julius Pfrommer} 
\author[1]{\\ Alexander Cebulla}
\author[4]{Simon Mangold}
\author[2]{Jürgen Beyerer}
\affil[1]{Institute for Anthropomatics and Robotics, Karlsruhe Institute of Technology, Germany}
\affil[2]{Fraunhofer Institute of Optronics, System Technologies and Image Exploitation IOSB, Germany}
\affil[3]{Fraunhofer Center for Machine Learning, Germany}
\affil[4]{wbk Institute of Production Science, Karlsruhe Institute of Technology, Germany \vspace{.5\baselineskip}}
\affil[ ]{\tt \small chengzhi.wu@kit.edu \quad xuelei.bi@student.kit.edu \quad julius.pfrommer@iosb.fraunhofer.de}
\affil[ ]{\tt \small alexander.cebulla@kit.edu \quad simon.mangold@kit.edu \quad juergen.beyerer@iosb.fraunhofer.de}

\maketitle
\thispagestyle{empty}

\begin{abstract}
On robotics computer vision tasks, generating and annotating large amounts of data from real-world for the use of deep learning-based approaches is often difficult or even impossible. A common strategy for solving this problem is to apply simulation-to-reality (sim2real) approaches with the help of simulated scenes. While the majority of current robotics vision sim2real work focuses on image data, we present an industrial application case that uses sim2real transfer learning for point cloud data. We provide insights on how to generate and process synthetic point cloud data in order to achieve better performance when the learned model is transferred to real-world data. The issue of imbalanced learning is investigated using multiple strategies. A novel patch-based attention network is proposed additionally to tackle this problem.
\end{abstract}

\section{Introduction}
\label{sec:intro}
Due to the rapid development of neural network algorithms, an increasing number of industrial companies and factories have started using deep learning (DL) methods for a variety of manufacturing and remanufacturing tasks in the past decade. In general, neural networks require a substantial amount of data in order to be trained, whereas for practical industrial applications, allocating and annotating a large amount of data is difficult or even impossible, especially when robots are involved. In the field of robotics, when the robot or manipulator directly interacts and samples with the real-world environment, there will be problems of low sampling efficiency and safety problems.

One possible solution to this problem is to apply simulation-to-reality (sim2real) method, which learns with simulated data and transfers the learned knowledge to real-world application. This is a common strategy used in robotics for learning robot movement controls \cite{Arndt2020MetaRL,Rajeswaran2018LearningCD} and robotic-related vision tasks \cite{Tobin2017DomainRF,Liu2020RealSimRealTF,Pashevich2019LearningTA}. In those computer vision tasks, simulated scenes are usually rendered into RGB images with possible auxiliary depth, thermal, or even flow images. Then, DL-based neural networks are pre-trained with the synthetic data and subsequently transferred to real-world use cases via domain adaption. However, most current sim2real work focuses on image data. Few researches apply sim2real methods on point cloud data. 
In this paper, we show a full pipeline of how to perform sim2real transfer learning on point clouds for a robotics use case as a part of a practical remanufactoring application. 

\begin{figure}[t]
    \centering
    \begin{subfigure}[b]{0.45\linewidth} 
        \centering
        \raisebox{0.3cm}{
        \includegraphics[width=\linewidth,trim=2 2 2 2,clip]{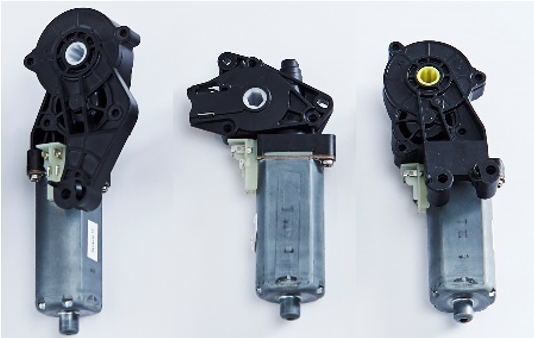}}
        \caption{}
        \label{fig:realMotor}
    \end{subfigure}
    \hspace{0.3cm}
    \begin{subfigure}[b]{0.45\linewidth} 
        \centering
        \includegraphics[width=\linewidth,trim=2 2 2 2,clip]{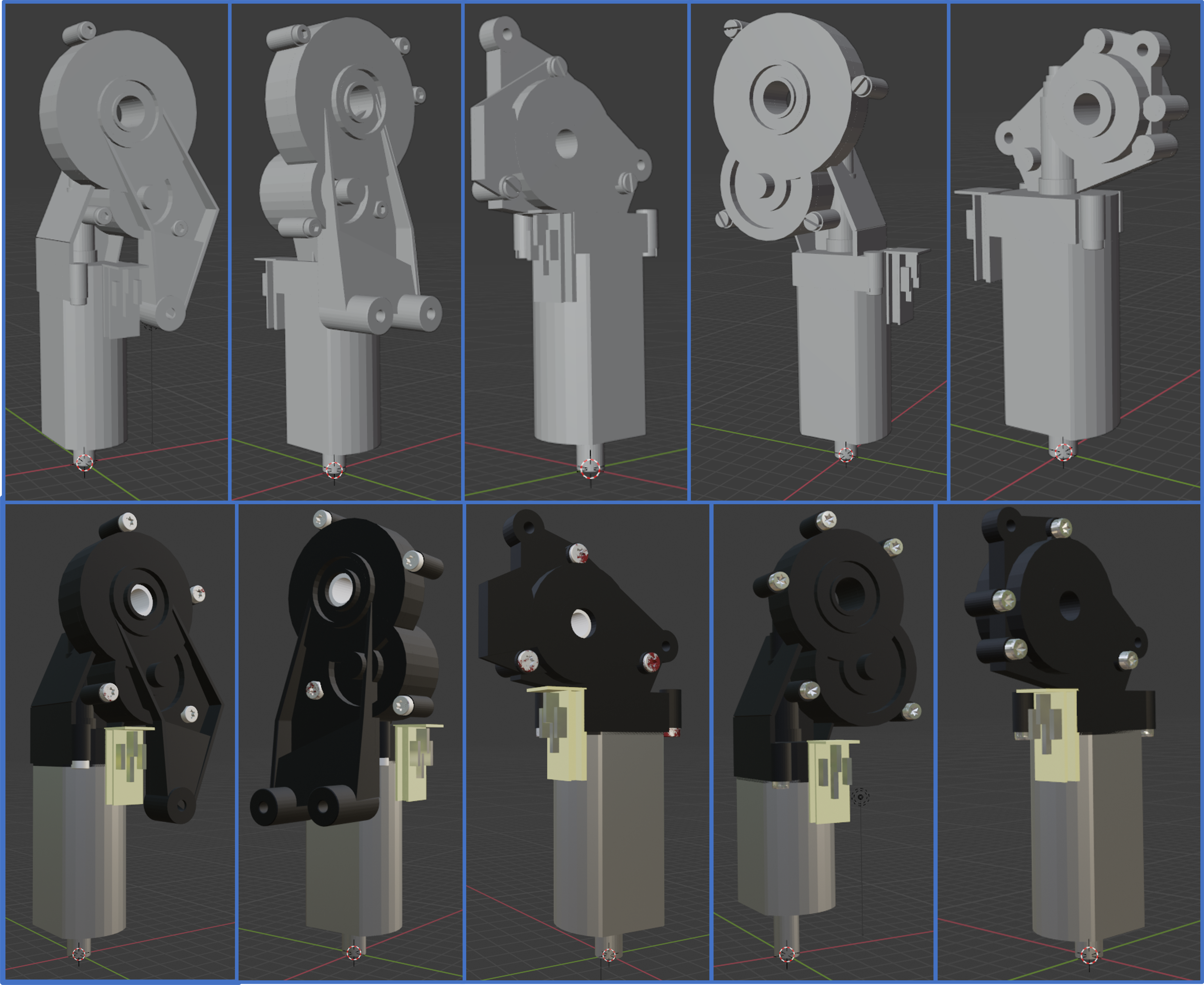}
        \caption{}
        \label{fig:motorDemo1}
    \end{subfigure}
    \begin{subfigure}[b]{0.7\linewidth} 
        \centering
        \includegraphics[width=\linewidth,trim=2 2 2 2,clip]{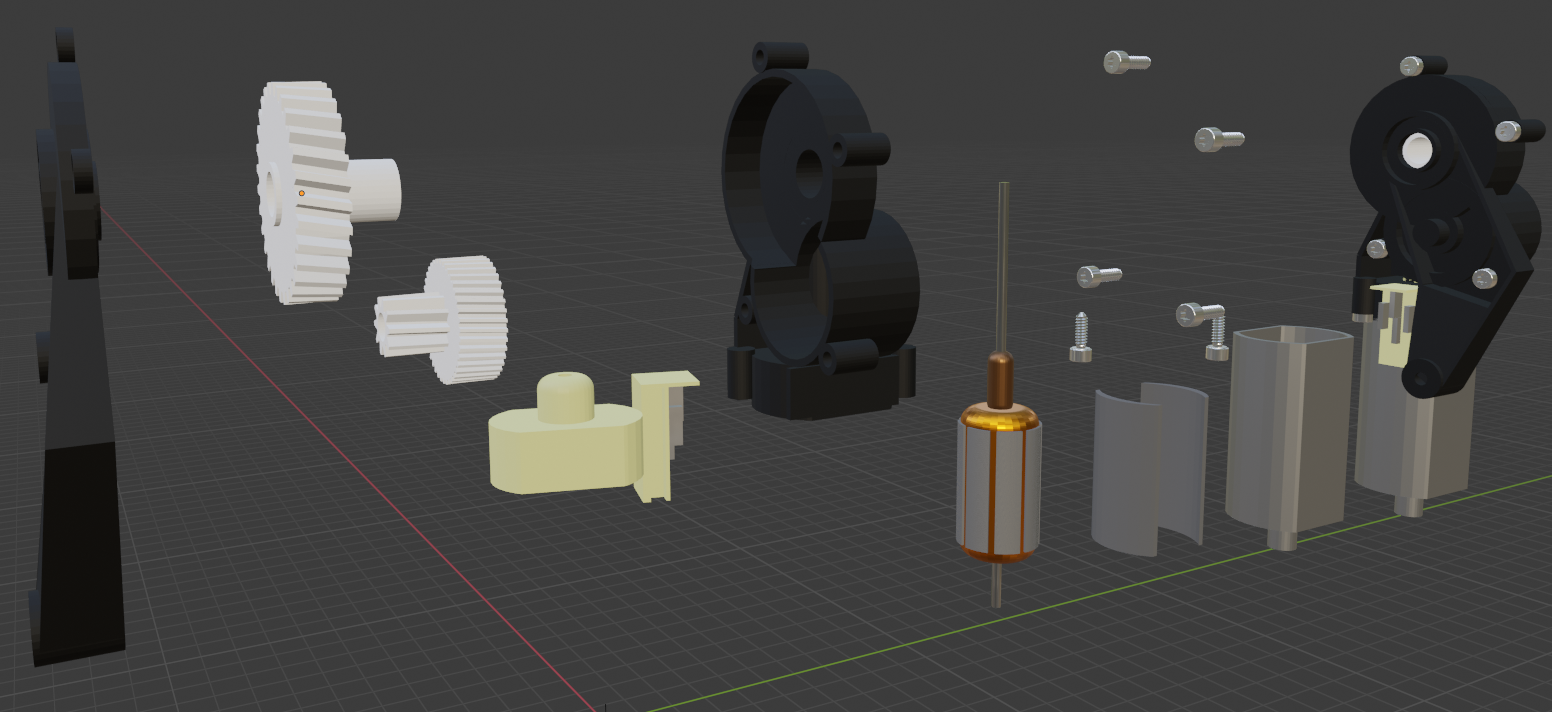}
        \caption{}
        \label{fig:motorDemo2}
    \end{subfigure}
    \caption{Real-world motors and generated demo motors. (b) Upper row: no textures added; bottom row: textures added and rendered. (c) An explosion figure of a generated motor. The original assembled motor model is also shown at the right most.  \vspace{-0.cm}}
    \label{fig:motorDemo}
\end{figure}

We consider the automated disassembly of different variants of actuators which are commonly used in vehicle manufacturing, \eg, as seat adjuster motors, window lift motors or rear door motors.
Several example motors are shown in Figure \ref{fig:realMotor}. The ultimate goal of this project is to use robots to perform automatic disassembly of motors, not only for the known motor types, but also for future variants with unseen specifications. In this case, generating a synthetic dataset with motor variants in simulated scenes for sim2real transfer learning \cite{Zhuang2021ACS} is a good solution. By learning the internal structure on part level, (\eg gear container, pole pot, electrical connection), processes on unseen variants which have similarities to the known population of actuators become feasible. This paper focuses on the first step of getting precise screw positions and orientations on motor covers for robots as one of the most important tasks for disassembly.


Generating a synthetic point cloud dataset for sim2real transfer learning has following advantages in our project: (i) a large synthetic dataset can be easily created, segmentation ground truth labels are given in the simulation, no manual annotation needed; (ii) motor variants with unseen specifications may be generated, which will strengthen the generalization ability of the trained network model; (iii) point cloud data contain richer 3D information for the learning. Using point cloud data avoids some problems that may occur when using image data, \eg, colors of the simulated images are far from realistic since it is hard to get the perfect textures for scene objects or to render the scene with perfect lighting conditions. When using the point cloud dataset, we use point coordinates information other than colors.


The remainder of this paper is structured as follows: Section \ref{sec:relatedWork} summarizes the state-of-the-art of 3D synthetic dataset creation, sim2real transfer learning, and point cloud segmentation. Section \ref{sec:dataset} shows a general pipeline of creating a synthetic dataset with simulated scenes. Section \ref{sec:generalPipeline} describes the whole sim2real learning framework and gives experimental results. Section \ref{sec:imbalancedLearning} additionally explores several strategies for imbalanced learning, including a novel patch-based attention network module. Finally, Section \ref{sec:conclusion} summarizes presented results and discusses future work.

\section{Related Work}
\label{sec:relatedWork}
\textbf{3D synthetic dataset.} 
Generating synthetic datasets as training data for machine learning purposes has already been widely discussed and used as a learning approach for various computer vision applications. 
In the past decade, many synthetic datasets of 3D models have been created, including the Princeton Shape Benchmark \cite{Shilane2004ThePS}, ModelNet \cite{Wu20153DSA}, ShapeNet \cite{Chang2015ShapeNetAI}, PartNet \cite{mo2018partnet}, etc. They collect large amounts of 3D models of different categories. A large dataset of 3D-printing models is provided in Thingi10K \cite{Zhou2016Thingi10KAD}, while a more recent ABC dataset \cite{Koch2019ABCAB} collects over 1 million CAD models including many mechanical components.
Regarding 3D scenes, \cite{Var17}, \cite{Ros16} and \cite{Kha19} generate synthetic datasets for the segmentation and detection of objects in virtual urban scenes. \cite{Hoa20} generates images from virtual garden scenes, while \cite{Trem18} creates a dataset for pose estimation. There are also works that generate synthetic point clouds. SynthCity \cite{Gri19} generates point clouds of urban scenes using Blender, while \cite{Pie19} also uses Blender but for the generation of point clouds of historical objects. 

\vspace{0.2cm}
\noindent \textbf{Sim2real transfer learning.} 
By allowing faster, more scalable, and lower-cost data collection than is possible in real-world, sim2real approaches show great impact on machine learning and have been applied in many fields including robotics and classic machine vision tasks. \cite{Tobin2017DomainRF}, \cite{Liu2020RealSimRealTF} and \cite{Vacaro2019SimtoRealIR} train neural network models on synthetic RGB images with domain randomization or domain adaption then transfer it to real-world, while Pachevish \etal \cite{Pashevich2019LearningTA} work with synthetic depth images. Also working with synthetic image data, Du \etal \cite{Du2021AutoTunedST} propose a method for automatically tuning simulator system parameters to match the real world. With the help of deep reinforcement learning \cite{Mnih2015HumanlevelCT}, robotics policies are directly used as training data for sim2real learning in some works \cite{Malmir2020RobustST,Arndt2020MetaRL}. A more detailed survey is given in \cite{Zhao2020SimtoRealTI}. Apart from robotics tasks, sim2real methods have also been widely used in other fields including autonomous driving \cite{Yue2019DomainRA,Revell2022Sim2realII}, medical diagnosis \cite{Abascal2021MaterialDI}, or even the control of atmospheric pressure plasma jets \cite{Witman2019SimtorealTR}.

\vspace{0.2cm}
\noindent \textbf{Point cloud segmentation.}
Before the appearance of PointNet \cite{Qi2017PointNetDL}, deep learning-based methods for point cloud segmentation are usually multi-view based \cite{lawin2017deep,boulch2017unstructured,audebert2016semantic,tatarchenko2018tangent} or volumetric-based \cite{maturana2015voxnet,jiang2018pointsift,le2018pointgrid}. PointNet \cite{Qi2017PointNetDL} is the first DL-based method that learns directly on points. It uses point-wise multi-layer perceptrons to extract global features. Its subsequent work of PointNet++ \cite{qi2017pointnet++} further considers local information. 
PointConv\cite{Wu2019PointConvDC} and KPConv \cite{thomas2019kpconv} propose point-wise convolution operators with which points are convoluted with neighbor points. Similar ideas are proposed in \cite{wang2018deep, hua2018pointwise}.
Simonovsky \etal \cite{simonovsky2017dynamic} takes each point as a graph vertex and applies graph convolution. In DGCNN \cite{wang2019dynamic}, EdgeConv blocks update the neighbor information dynamically. RandLA-Net \cite{Hu2020RandLANetES} learns attention scores for points as a soft mask to replace the original pooling layer. GAPNet \cite{Chen2021GAPNetGA} and Liang \etal \cite{Liang20203DIE} propose graph-attention operations with neighbor points to learn coefficients. More recently, transformer-based methods are starting to trend. PCT \cite{Guo2021PCTPC} pioneers on this direction by replacing the encoder layers in the original PointNet \cite{Qi2017PointNetDL} framework with self-attention layers, while PT \cite{Zhao2020PointT} is based on U-Net \cite{Ronneberger2015UNetCN}. SortNet is proposed in \cite{Engel2021PointT} to learn sub-point clouds, with which attention operations are applied on their latent features and the global feature to perform local-global attention.

\begin{figure}[t]
    \centering
    \begin{subfigure}[b]{0.45\linewidth} 
        \centering
        \includegraphics[width=\linewidth,trim=2 2 2 2,clip]{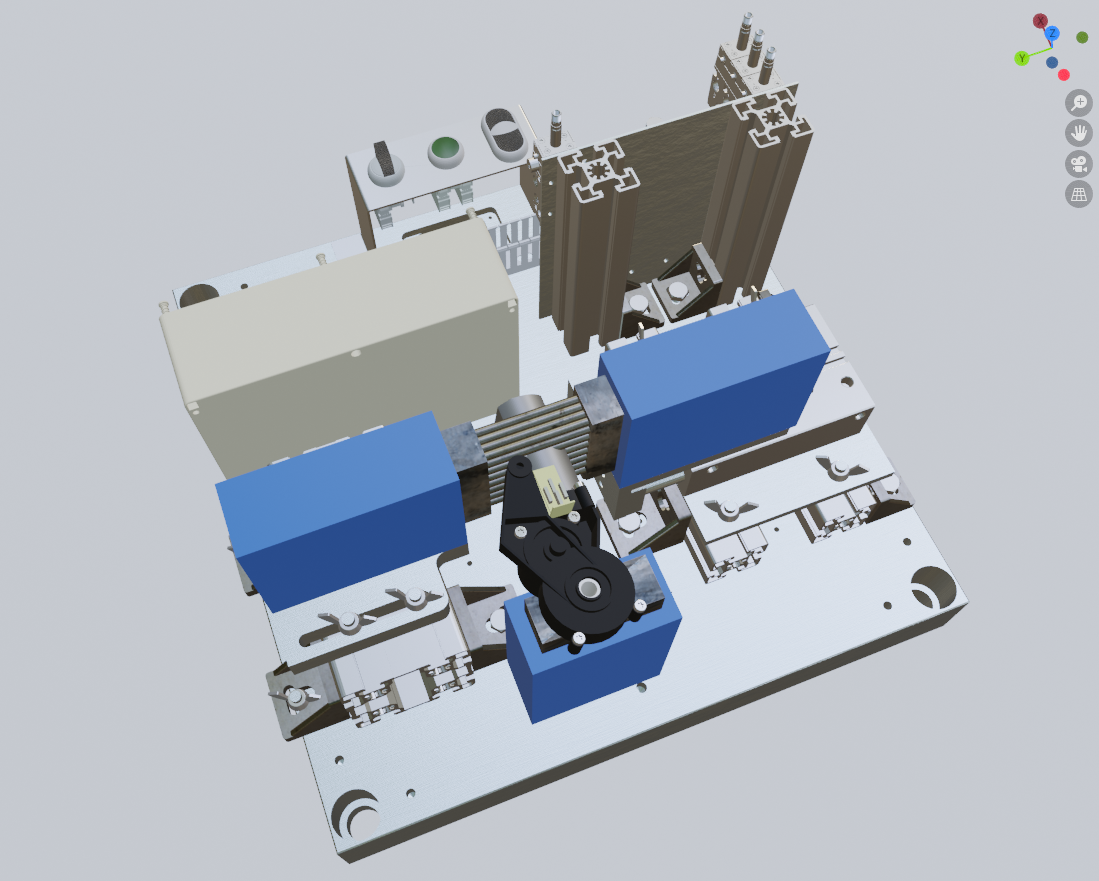}
        \caption{}
        \label{fig:blenderScene} \vspace{0.2cm}
    \end{subfigure}
    \hspace{0.2cm}
    \begin{subfigure}[b]{0.48\linewidth} 
        \centering
        \includegraphics[width=\linewidth,trim=2 2 2 2,clip]{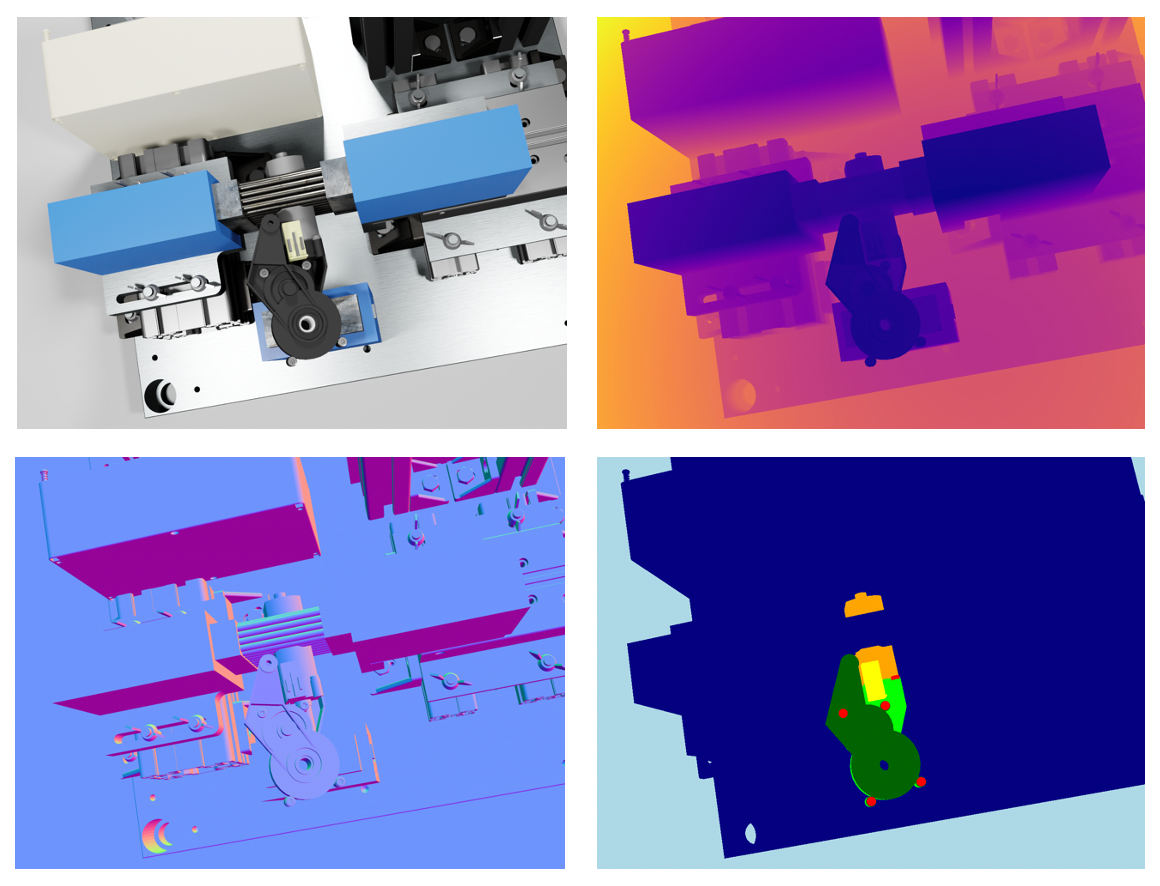}
        \caption{}
        \label{fig:demoImage} \vspace{0.2cm}
    \end{subfigure}
    \begin{subfigure}[b]{0.85\linewidth} 
        \centering
        \includegraphics[width=\linewidth,trim=2 2 2 2,clip]{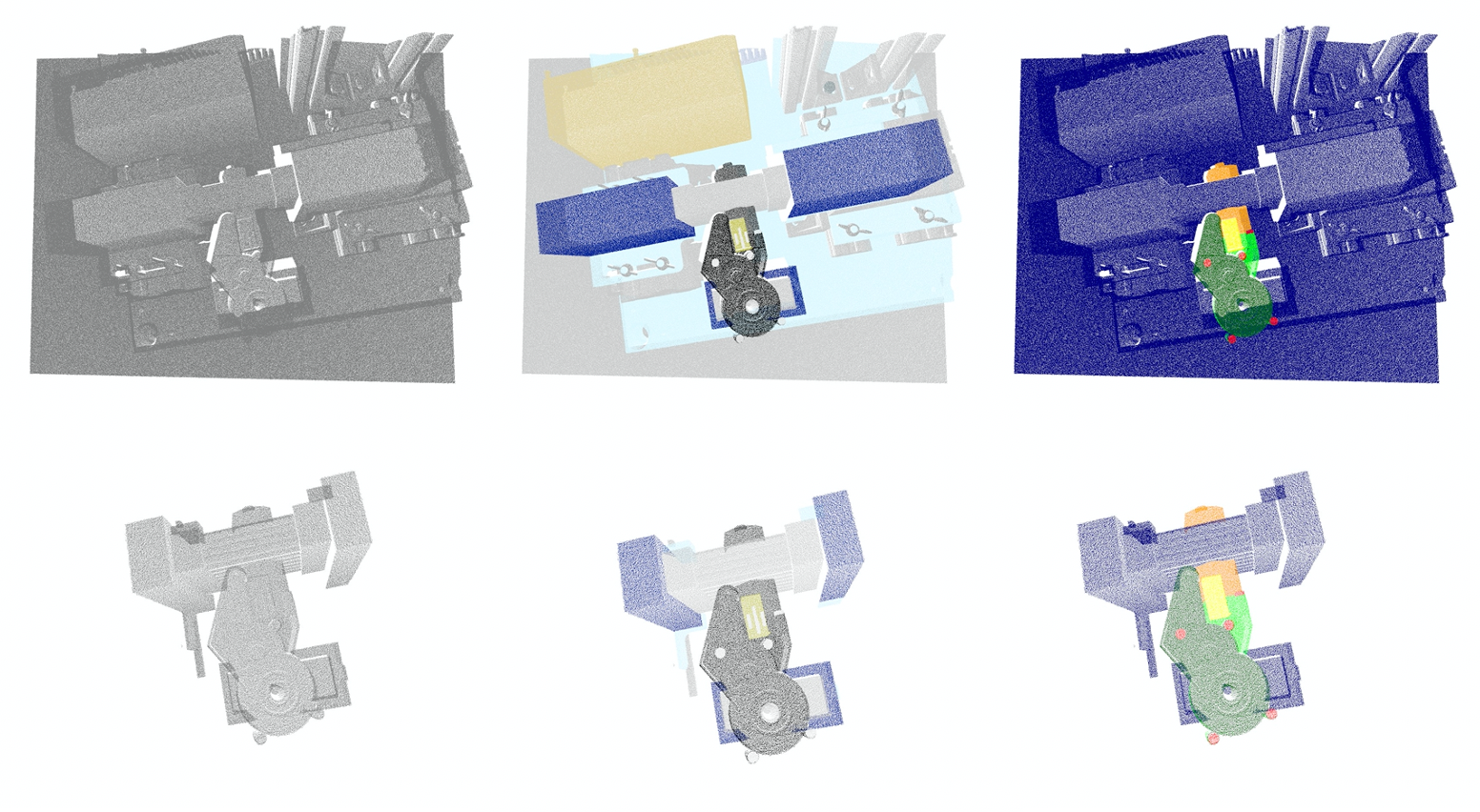}
        \caption{}
        \label{fig:demoPC} 
    \end{subfigure}
    \caption{Synthetic dataset generation: (a) simulated scene built in Blender; (b) synthetic image data generated with BlenderProc; (c) synthetic point cloud data generated with BlenSor.}
    \label{fig:dataset}
\end{figure}

\section{Synthetic Dataset Generation}
\label{sec:dataset}
\subsection{Synthetic Mesh Model Generation} 
To easily generate motor mesh models of a variety of specifications, we create a Blender addon based on the motor types we have. As an open source software, Blender \cite{bonatti2016blender} is a proven tool that performs well in modeling shapes and creating highly customizable addons. Our addon is able to generate motor mesh models with various specifications and save them in desired file formats. Each component of a generated motor can also be saved separately. 

The generated models contain the following components: (i) Pole Pot; (ii) Electric Connection; (iii) Gear Container; (iv) Cover and (v) Screws. Those are the five main categories we need perform segmentation on. Additionally, following inner components have also been generated : (vi) Magnets; (vii) Armature; (viii) Lower Gear and (ix) Upper Gear as presented in \ref{fig:motorDemo2}. However, since we only focus on the first step of unscrewing process in this paper, inner parts will not be investigated. 
To generate motors with various specifications, we provide lots of parameter options that control the type, size, position and rotation of different parts of motor, \eg screw position, gear size, or pole pot length.
Figure \ref{fig:motorDemo1} shows ten generated demo motors with different parameters and an exploded view of a demo motor. All the individual components mentioned above are modeled separately as illustrated.

\subsection{Synthetic Point Cloud Generation} 
The generated mesh models are further used to create synthetic image and point cloud datasets. A simulated scene is built in Blender for it. Apart from the lights and cameras, the Blender scene also contains the model of the real-world clamping system and a background panel. The camera rotates randomly on top of the scene within a certain range yet always towards the motor. To create image dataset, apart from the scene images rendered by Blender directly, BlenderProc \cite{denninger2019blenderproc} can be used to generate corresponding depth images, normal images, and segmentation ground truth images as shown in Figure \ref{fig:demoImage}. Or in our case, BlenSor \cite{Gschwandtner2011BlenSorBS} is used to simulate the sensors to create point cloud dataset. 
Figure \ref{fig:demoPC} gives a demo of generated point cloud colored in the material color or its segmentation ground truth. 

\begin{figure*}[t]
    \centering
    \includegraphics[width=0.98\linewidth,trim=2 2 2 2,clip]{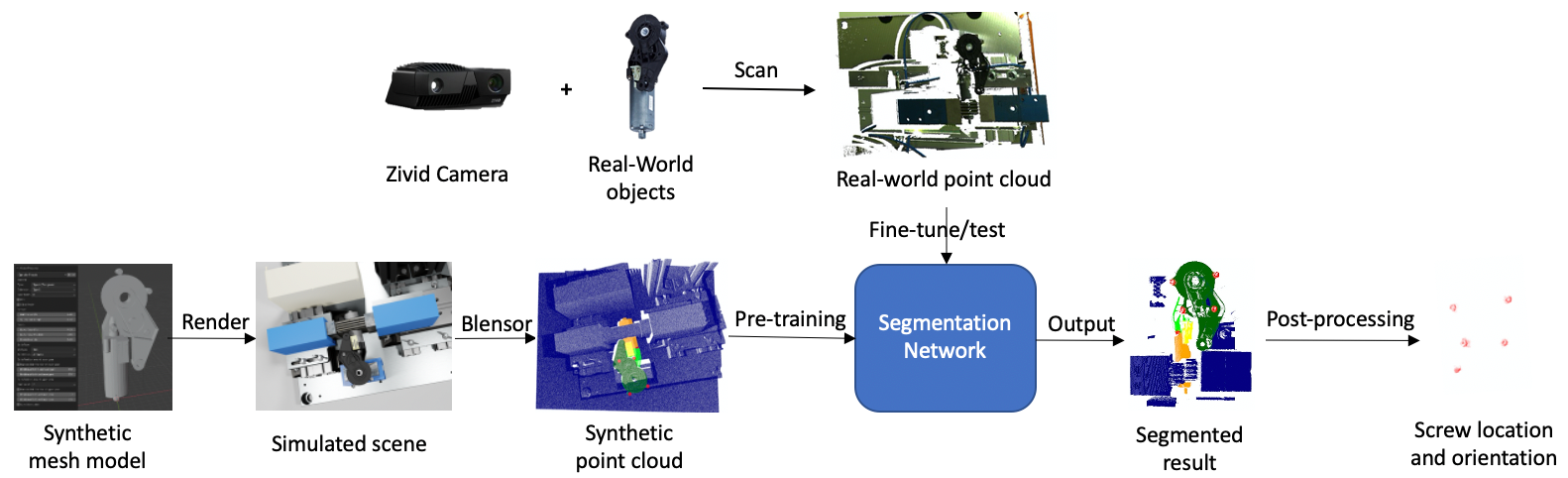}
    \caption{Sim2real transfer learning piepline for the point cloud segmentation task in our industrial application.}
    \label{fig:framework}
\end{figure*}

\subsection{Data Pre-processing and Augmentation} 
\label{sec:preprocessing}
The generated point clouds are of relatively large size. Each point cloud contains around 1.2 million points. 
However, neural networks have a limitation of point number per batch, common choices are 1024/2048/4096 points. Direct downsampling makes the sub-point clouds contain too less points for tail categories (\eg, screws in our case) thus does not work.
To deal with large point clouds, common industrial applications use sliding voxels to voxelize the point cloud space and perform prediction voxel-wise. In our case, we want to perform the point cloud segmentation fast and precisely to prevent the robot from being idle for a too long time. 
Using the prior knowledge that motors are always clamped in a relatively fixed position in the clamping system and only the area around the motor is of our interest, we restrict the sampling region by cropping a cuboid area around that location and use it as input to the segmentation neural network. All other residual points are labeled as background points directly. The size of the cropped point cuboid is of only around $10\%$ of the raw point cloud. By doing so, direct downsampling the point cuboid into sub-point clouds of 2048 points makes the tail categories still have enough points for learning. A cropped point cuboid demo is given in Figure \ref{fig:demoPC}.

Apart from pre-processing, augmentation is also an important part to improve the generalization ability of the models. Common augmentation methods include random rotation and random jittering over the whole point cloud. In our case, we additionally introduce other augmentation methods of (i) random cuboid size along all three axes; (ii) random mild translation and rotation of motors; and (iii) adding random size and random position hovering tiles over the clamping system as scene masks, which is similar to the masking augmentation on images. An detailed ablation study regarding the augmentations is given in Section \ref{sec:aug}.

\section{Sim2real Point Cloud Learning}
\label{sec:generalPipeline}
As illustrated in Figure \ref{fig:framework}, apart from the synthetic point cloud dataset generation, the whole pipeline consists following other steps: pre-train the network model on synthetic data, fine-tune the network model on real-world data, and post-processing for screw information.

\subsection{Pre-training on Simulated Scenes}
As reviewed in Section \ref{sec:relatedWork}, there are a variety of neural network models that can be used for the point cloud segmentation task. 
Since the backbone itself is not of our main focus, balancing the performance and the computation time, we use DGCNN \cite{wang2019dynamic} as our backbone network for the segmentation task. To better deal with the camera perspective variance, a spatial transform network (STN) \cite{Qi2017PointNetDL} is introduced at the beginning.

In the previous step, a dataset of 1000 random scenes with 1000 random motors is generated. We use $80\%$ of it for training and $20\%$ for test. In the pre-training process, the training takes 100 epochs and the batch size is 16. We use an initial learning rate of 0.01, and a $cos$ decay scheduler with a final learning rate of 0.00001. The optimization method is stochastic gradient descent (SGD). We use the widely used cross entropy for the segmentation loss $L_\text{seg}$. An auxiliary STN rotation loss $L_\text{rot}$ which describes the L2 difference between the learned rotation matrix and the ground truth (saved during the augmentation) is additionally applied with a small weight $\alpha$. The total loss is defined as:
\vspace{-0.2cm}
\begin{equation}
    L_\text{total} = L_\text{seg} + \alpha L_\text{rot} \vspace{-0.2cm}
\end{equation}
We set $\alpha = 0.01$. During the training, the common segmentation metric, mean intersection over union (mIoU), is used to measure the model performance. The IoU of screw category is used as an additional metric since screws are of our main focus.
During both pre-training and fine-tuning processes, we save the model that performs best on the screw IoU metric.

\begin{table}[t]
\caption{Numerical results of models with different settings. The second column indicates whether the model is pre-trained on the synthetic dataset or not. Both results in the pre-training step (Simulation) and the fine-tuning step (Reality) are given. Note that the results in Simulation columns are the test results on synthetic test dataset, other than real-world test dataset. Same below.}
\centering
\resizebox{1\linewidth}{!}{
\begin{tabular}{cccccccc}
\hline
\multirow{2}{*}{STN} & \multirow{2}{*}{Pre-train} & \multirow{2}{*}{} & \multicolumn{2}{c}{Simulation} & \multirow{2}{*}{} & \multicolumn{2}{c}{Reality} \\ \cline{4-5} \cline{7-8} 
 &  &  & mIoU & screw IoU &  & mIoU & screw IoU \\ \hline
- & - &  & - & - &  &0.8707  & 0.4884\\
\checkmark & - &  & - & - &  & 0.9030 &0.6272  \\
- & \checkmark &  & 0.9202 & 0.7120 &  & 0.9187  & 0.6830 \\
\checkmark & \checkmark & & \textbf{0.9675} & \textbf{0.8875} &  & \textbf{0.9375}  & \textbf{0.7842}  \\ \hline
\end{tabular}}
\label{table:stn}
\end{table}

\begin{figure*}[t]
    \centering
    \begin{tabular}{p{2.0cm}<{\centering} p{2.7cm}<{\centering} p{3.2cm}<{\centering} p{3.2cm}<{\centering} p{3.2cm}<{\centering}}
    input & GT & w/o pre-train & only pre-train & fine-tuned \\ \hline
    \end{tabular}
    \includegraphics[width=0.96\linewidth,trim=2 2 2 2,clip]{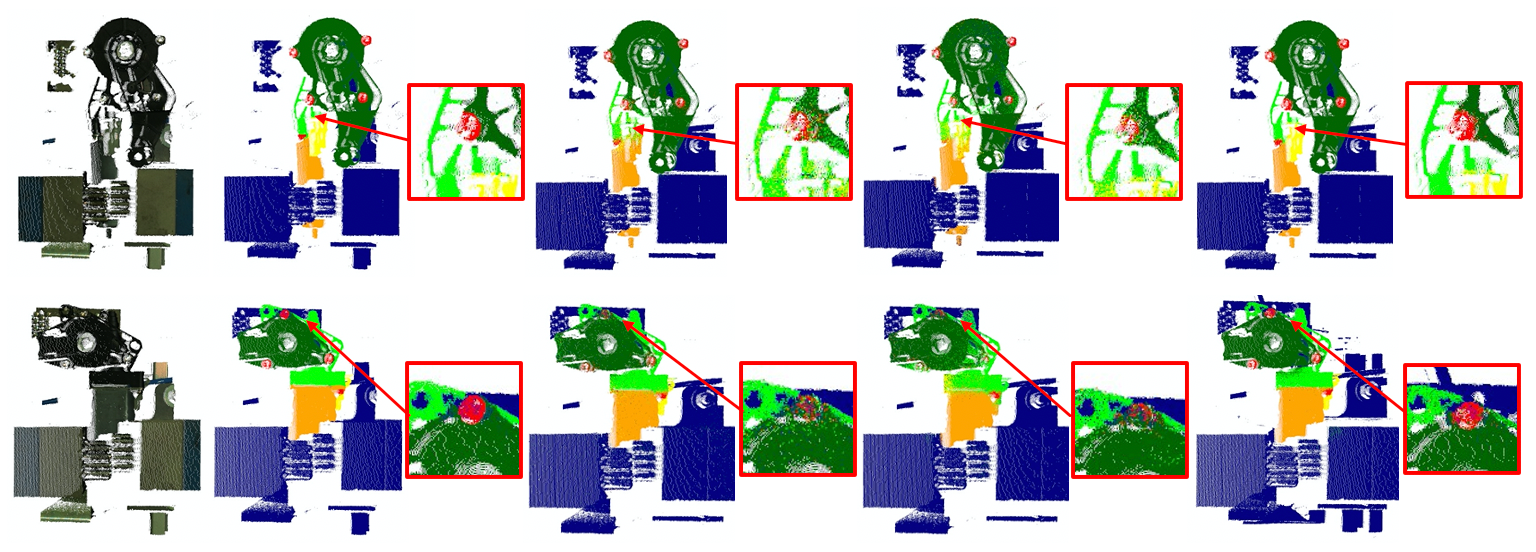}
    \caption{Segmentation result visualization on real-world test data. Without pre-train means the network is trained on real-world data directly. Only pre-train means the network is trained only on synthetic dataset and directly used for the testing on real-world data. Fine-tuned means the network is both pre-trained and fine-tuned.}
    \label{fig:segVis}
\end{figure*}

\begin{table*}[h]
\caption{Ablation study on augmentation methods. Aug 1: random rotation and random jittering over the cuboid point cloud. Aug 2: random cuboid size along all three axes. Aug 3: random mild translation and rotation of motors. Aug 4: adding random position hovering tiles over the clamping system as scene masks.}
\centering
\begin{tabular}{ccccccccccc}
\hline
\multirow{2}{*}{Dataset} & \multicolumn{4}{c}{Augmentations} & \multirow{2}{*}{} & \multicolumn{2}{c}{Simulation} & \multirow{2}{*}{} & \multicolumn{2}{c}{Reality} \\ \cline{2-5} \cline{7-8} \cline{10-11} 
 & aug 1 & aug 2 & aug 3 & aug 4 &  & mIoU & screw IoU &  & mIoU & screw IoU \\ \hline
dataset 1 &\checkmark & &  &  &  &  0.9751&  0.9141&  &  0.9280&0.7518   \\
dataset 2 &\checkmark &\checkmark &  &  &  & \textbf{0.9801} &  \textbf{0.9372} &  &  0.9368& 0.7799 \\ 
dataset 3 &\checkmark &\checkmark & \checkmark &  &  & 0.9742 &0.9161  &  & 0.9370 & 0.7815  \\ 
dataset 4 &\checkmark &\checkmark &\checkmark &\checkmark &  &  0.9675 & 0.8875 &  &  \textbf{0.9375} & \textbf{0.7842}  \\ 
\hline
\end{tabular}
\label{table:aug}
\end{table*}

\subsection{Fine-tuning on Real-world Scenes}
During the fine-tuning step, we load the network parameters from the pre-training step and perform transfer learning, \ie, adapting the network model from simulated scenes to real-world scenes. We took 26 real-world point clouds with Zivid camera and manually labeled them. 20 scenes are used for fine-tuning and 6 scenes are used for test.
Note that this is not a small dataset since the point cloud cuboid from each scene contains around 200,000 points and can be sampled to around 100 sub-point clouds of 2048 points.
In the fine-tuning process, the training takes 300 epochs and the batch size is 16. We use an initial learning rate of 0.001, and a $cos$ decay scheduler with a final learning rate of 0.00001. The optimizer, loss, evaluation metrics are identical to that in the pre-training step.

Note that after the model is trained, the dataloader for the test process is different from that for the training process. When training the network, each input point cuboid splits into sub-point clouds of 2048 points as much as it can, and the residual points are discarded. When testing the trained model, the residual points are not discarded but completed into a sub-point cloud of 2048 points with other random points resampled from the original input.

\subsection{Quantitative and Qualitative Results}
Numerical results over the metrics of mIoU and screw IoU are given in Table \ref{table:stn}. From it, we can observe that using the STN module improves the performance drastically. On the other hand, using the sim2real transfer learning with two steps of pre-training and fine-tuning also boosts the performance. Combining both gets even better performance.

Some qualitative results are given in Figure \ref{fig:segVis}. From it, we can observe that direct training on the real-world data performs decent on most points but not so good on tail categories. With pre-training on the simulated scenes and fine-tune the model on real-world scenes can achieve a much better segmentation result, especially for the tail categories.


\begin{figure*}[h]
    \centering
    \includegraphics[width=1\linewidth,trim=2 2 2 2,clip]{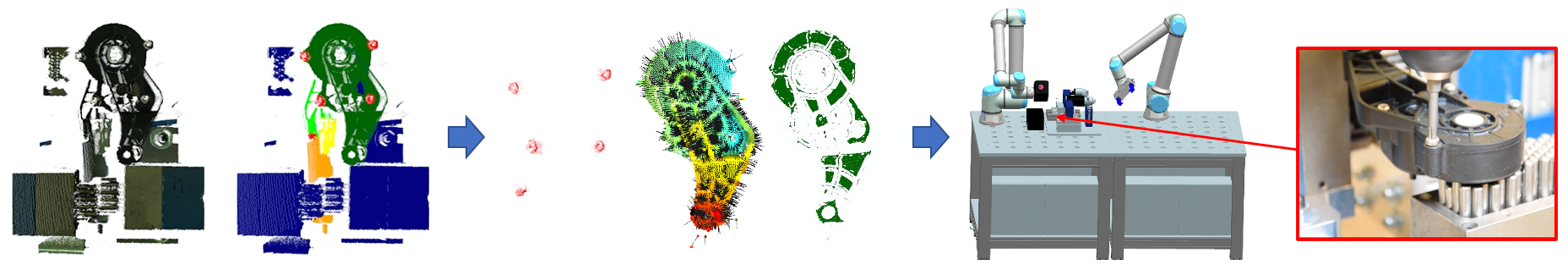}
    \caption{Post-processing for screw information.}
    \label{fig:post}
\end{figure*}

\begin{figure*}[t]
    \centering
    \begin{subfigure}[b]{0.45\linewidth}
        \centering
        \includegraphics[width=\linewidth,trim=2 2 2 2,clip]{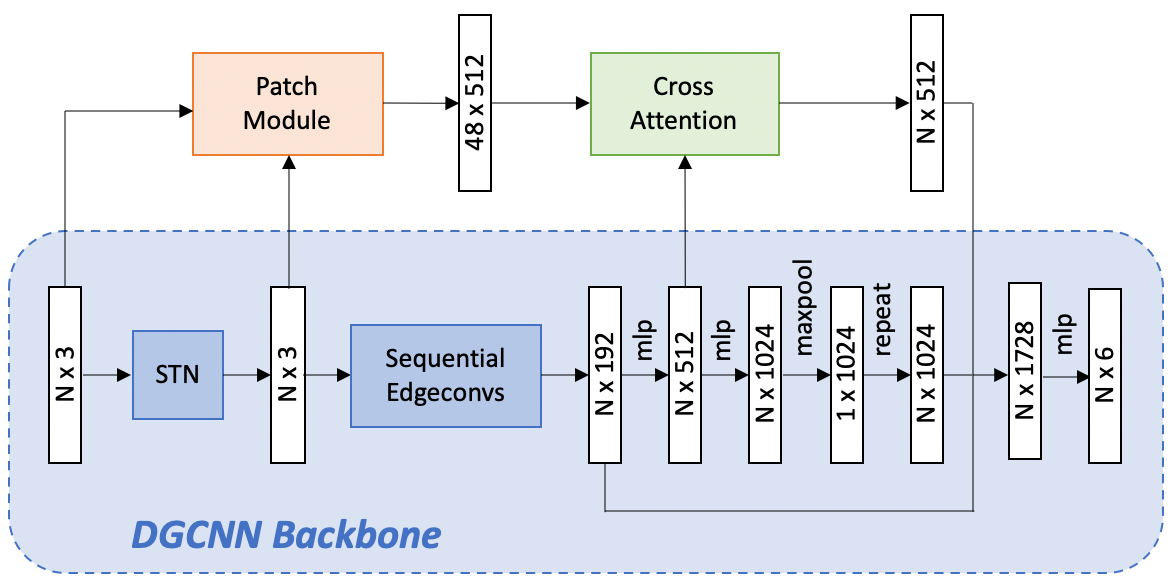}
        \caption{\vspace{-0.3cm}}
        \label{fig:pacthFull} 
    \end{subfigure}
    \hspace{0.02cm}
    \begin{subfigure}[b]{0.26\linewidth}
        \centering
        \includegraphics[width=\linewidth,trim=2 2 2 2,clip]{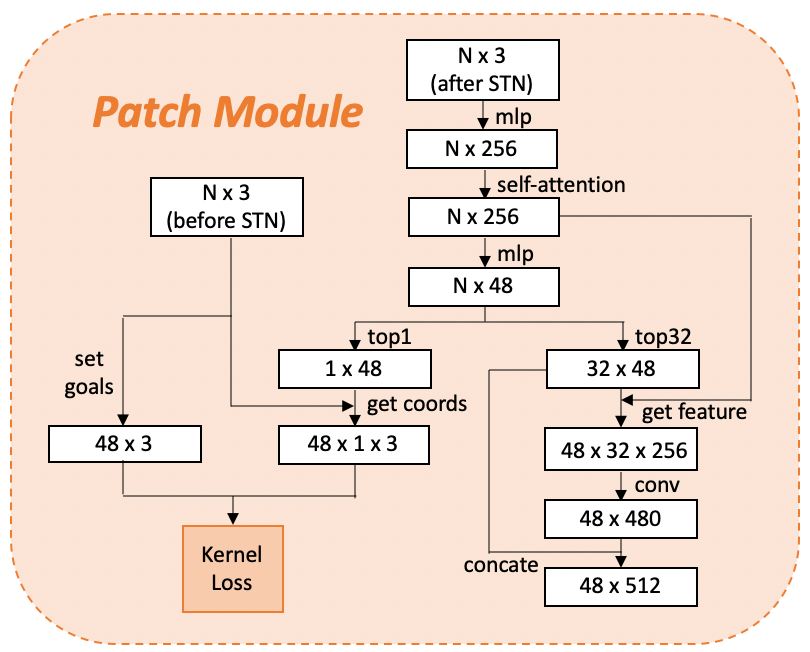}
        \caption{\vspace{-0.3cm}}
        \label{fig:patchModule} 
    \end{subfigure}
    \hspace{0.02cm}
    \begin{subfigure}[b]{0.235\linewidth}
        \centering
        \includegraphics[width=\linewidth,trim=2 2 2 2,clip]{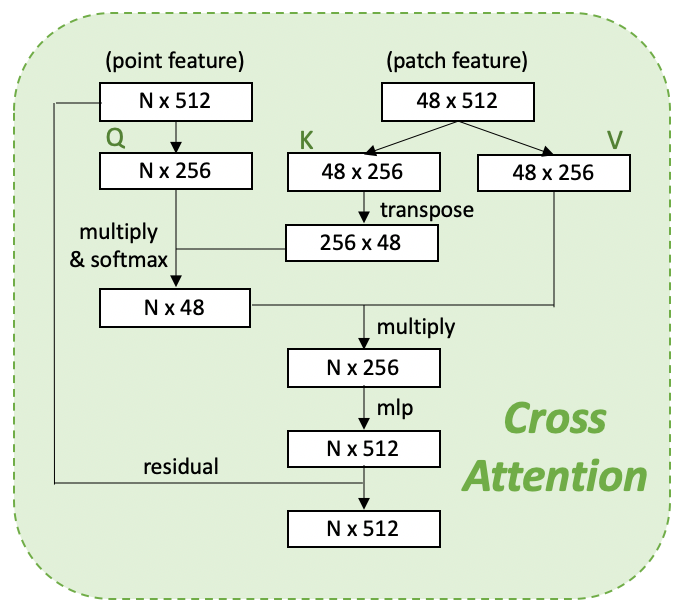}
        \caption{\vspace{-0.3cm}}
        \label{fig:crossAtten} 
    \end{subfigure}
    \caption{Patch-based attention network. (a) Full architecture. (b) Proposed patch module, has an additional defined kernel loss, outputs patch feature. (c) Cross attention module.  \vspace{-0.cm}}
    \label{fig:patchNet}
\end{figure*}

\subsection{Post-processing for Screw Information} 
After the network model is fine-tuned, given a real-world point cloud as input, the network outputs the segmentation result. The post-processing step aims at getting screw locations and orientations with the screw points that have been segmented out. To get screw locations, clustering algorithms are firstly used to group segmented screw points. In our case, we use the 
Density-Based Spatial Clustering of Applications with Noise (DBSCAN) \cite{Ester1996ADA, schubert2017dbscan} algorithm for clustering. With appropriate parameter settings, each cluster is one screw. Using the prior knowledge that the bottom most screw is always the side screw, all the clusters above are cover screws that need to be unscrewed in this process. The cover screw locations are obtained by computing the center of each cluster. 
For the screw orientations, processing on the screw points directly is problematic since screws have uneven surfaces. Using the prior knowledge that all cover screws are having the same orientation as the flat cover, the screw orientation is actually identical to the normal of the cover flat part in our case. We hence apply DBSCAN on the points that are segmented as the cover category but with their estimated normals other than coordinates. With appropriate parameter settings, we can make the cluster number to be only one, which means all the points whose normal vectors are similar to the cover flat part are clustered together. Then the cover normal, or the screw orientation, is obtained by averaging all the normal vectors of those points. The process is illustrated in Figure \ref{fig:post}.

\subsection{Ablation Study on Data Augmentation}
\label{sec:aug}
A variety of augmentation methods have been used on our synthetic dataset. To validate their effectiveness, an ablation study is performed by generating several different datasets and using a same network architecture for the pre-training and fine-tuning processes. The numerical results are given in Table \ref{table:aug}. In the used four augmentation methods, the former two methods are augmentations performed during the data pre-processing, and the latter two methods are augmentations performed during the data generation.
From Table \ref{table:aug}, we can observe that randomly changing the cuboid size improves the segmentation performance on both pre-training and fine-tuning steps. On the other hand, while the latter two augmentation methods decreases the pre-training performance, they both improves the fine-tuning performance when the model is transferred to real-world point cloud data. In this paper, we use dataset 4 for most other experiments.

\section{Imbalanced Learning}
\label{sec:imbalancedLearning}
Imbalanced learning, also referred as long-tail learning in the classification tasks, is a problem where the distribution of examples across the known categories is biased or skewed. In our case, original synthetic point clouds are mostly occupied with background points (around $96\%$) and have extremely less screw points (around $0.1\%$). After applying the \textbf{sample region restriction}, \ie, the cuboid crop strategy as illustrated in Figure \ref{fig:demoPC} and described in subsection \ref{sec:preprocessing}, in each point cuboid, background points take up around $64\%$ while screw points take up around $1\%$. The imbalanced problem has been alleviated. Meanwhile, it is still worth investigating to further improve the performance. 
Several additional strategies are proposed to deal with the imbalanced learning problem in this paper. They are proposed from the perspectives of data augmentation, weighting loss, and extra network block respectively.

\begin{table*}[h]
\caption{Ablation study on proposed strategies for imbalanced learning. \vspace{-0.cm}}
\centering
\begin{tabular}{cccccccccc}
\hline
\multirow{2}{*}{\begin{tabular}[c]{@{}c@{}}Sample region\\ restriction\end{tabular}} &
\multirow{2}{*}{\begin{tabular}[c]{@{}c@{}}Focused\\ sampling\end{tabular}} & \multirow{2}{*}{\begin{tabular}[c]{@{}c@{}}Weighting\\ loss\end{tabular}} & \multirow{2}{*}{\begin{tabular}[c]{@{}c@{}}Patch-based\\ attention\end{tabular}} & \multirow{2}{*}{} & \multicolumn{2}{c}{Simulation} & \multirow{2}{*}{} & \multicolumn{2}{c}{Reality} \\ \cline{6-7} \cline{9-10} 
&  &  &  &  & mIoU & screw IoU &  & mIoU & screw IoU \\ \hline
\checkmark & - & - & - &  & 0.9675 & 0.8875 &  & 0.9375 & 0.7842 \\
\checkmark & \checkmark & - & - &  & 0.9627 & 0.8661 &  & 0.9376 & 0.7570 \\
\checkmark & - & \checkmark & - &  & 0.9668 & 0.8862 &  & 0.9409 & 0.7717 \\
\checkmark & - & - & \checkmark &  & 0.9693 & 0.9063 &  & \textbf{0.9462} & \textbf{0.7968} \\
\checkmark & \checkmark & - & \checkmark &  & 0.9644 & 0.8850 &  & 0.9389 & 0.7622 \\
\checkmark & - & \checkmark & \checkmark &  & \textbf{0.9729} & \textbf{0.9165} &  & 0.9412 & 0.7794 \\ 
\checkmark & \checkmark & \checkmark & \checkmark &  & 0.9680 & 0.8972 &  & 0.9401 & 0.7683 \\ \hline
\end{tabular}
\label{table:imbalanced}
\end{table*}

\begin{figure*}[h]
    \centering
    \includegraphics[width=1\linewidth,trim=2 2 2 2,clip]{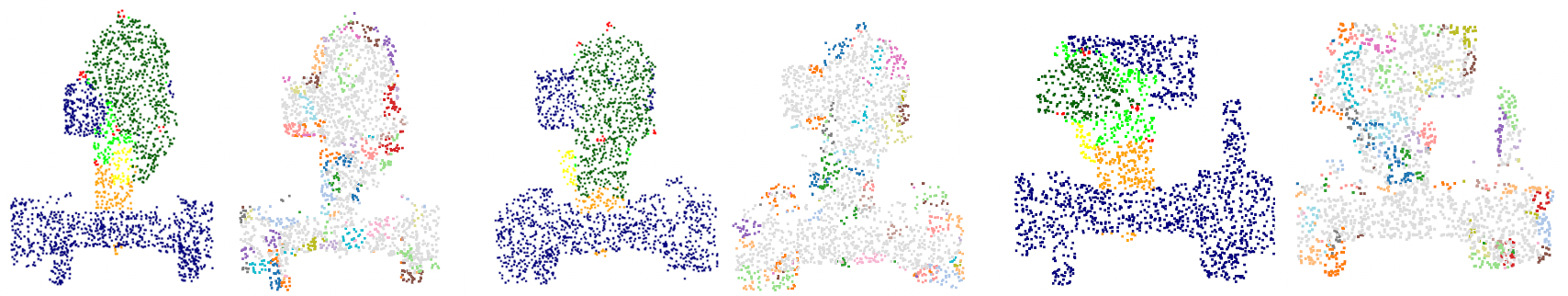}
    \caption{Visualizing learned patches on sub-point clouds of 2048 points.}
    \label{fig:patchVis}
\end{figure*}

\subsection{Focused Sampling for Tail Categories}
One common strategy to deal with the long-tail problem in classification tasks is resampling like manually adding samples of the tail categories. In our case, the key category of screw only occupies a extreme small portion of the whole point cloud compared to other components, hence it is possible to increase the number of screw points by densifying them. The strategy is as follows. For each point $p_1$ belongs to the screw category, get its nearest same category point $p_2$. (a) If the nearest same category point of $p_2$ is also $p_1$, this means both points are not at the cluster boundary, hence a new point is added with the coordinate of $p_\text{add} = p_1 + \frac{1}{3} (p_2 - p_1)$. (b) If the nearest same category point of $p_2$ is not $p_1$, this means $p_1$ is likely to be an outlier point or at the cluster boundary, hence a new point is added with the coordinate of $p_\text{add} = p_1 + \frac{2}{3} (p_2 - p_1)$. Above operation doubles the point number of the tail category in the training dataset.

\subsection{Weighting Category Loss}
Adding additional weights to each category when computing the cross entropy loss is another widely used strategy. Most current DL packages provide such an optional argument in their in-built loss functions. However, it is an unsolved question that what is the best way to compute and set the category weights. In this paper, we propose a following method. Assume the point cloud has $M$ categories and $N$ points in total. For each category that has $n_i (i=1,2,\dots,M)$ points, its ratio is given as $r_i = n_i / N$. Then a scaled ratio $sr_i$ is computed by decreasing the original ratio difference with a cubic root operation as $sr_i = t_i / (\sum_{i=1}^{M} t_i)$, where $t_i = (max(r_1, r_2, \dots, r_M)/r_i)^{\frac{1}{3}}$. In this case, smaller $r_i$ means the corresponding category gets a larger $sr_i$.  
However, using $sr_i$ directly leads to a huge decrease on the loss magnitude. To eliminate the possible problem caused by it, an additional factor is computed as $f = \sum_{i=1}^{M} r_i \times sr_i$ and multiplied. Hence the final category weight is given as $\omega_i = sr_i \times f$ for each category.

\subsection{Patch-based Attention Network}
Additionally, we propose a novel patch-based attention network to deal with the imbalanced learning problem. The key idea is to force the network to learn a same number of kernel points for all categories by using an additional kernel loss. In our task, we have 6 categories in total and we select 8 kernel points per category. For each category, the ground truth kernel points are obtained by performing K-means algorithm on all points of this category. After 8 clusters are grouped, cluster centers are computed and their nearest neighbor points that belong to this category are defined as kernel points. The kernel loss is a L2 loss between the goal kernels and learned kernels. Hence the total loss in this case is defined as:
\vspace{-0.2cm}
\begin{equation}
    L_\text{total} = L_\text{seg} + \alpha L_\text{rot} + \beta L_\text{ker} \vspace{-0.2cm}
\end{equation}
where $\beta$ is a loss weight. We set $\beta = 0.05$. Apart from the obtained kernel points, their 32 neighbor points are grouped to form a patch and a convolution layer is used to get patch-wise features. The patch-wise features are further used to perform cross attention with the point-wise features learned from the DGCNN backbone (lower branch in Figure \ref{fig:pacthFull}). This is a patch-to-point cross attention, \ie, using patch features to represent point features. The output is concatenated back to the lower branch for final segmentation.
Detailed network designs are given in Figure \ref{fig:patchNet}.

\subsection{Experimental results} 
All the experiments are conducted with the same dataset on which all augmentation methods are applied (dataset 4 in Table \ref{table:aug}).
Numerical results of them are presented in Table \ref{table:imbalanced}. From it, we can observe that the focused sampling strategy always leads to a worse performance. One possible reason is that this operation is only performed during training. For test cases, labels are segmentation goals and are not provided hence the focused sampling operation is not applicable. This causes data distribution difference between the training set and the test set thus leads to bad performance. 
The loss weighting strategy improves the performance on synthetic data but degrades the performance on real-world data slightly. This indicates the strategy contributes to a better pre-training yet not performing well on transfer learning. On the other hand, our proposed patch-based attention module improves the performance on both steps. To give better insights of our proposed module, learned patches of some sub-point clouds are visualized in Figure \ref{fig:patchVis}. It shows that our method forces the network to learn patches around boundaries or other informative places.

\section{Conclusion}
\label{sec:conclusion}
In this paper, we adopt sim2real transfer learning method for an industrial application on point cloud data. Following the pipeline, synthetic dataset are generated in simulated scenes. The network model is firstly pre-trained on the synthetic data and then fine-tuned on the real-world data. Both quantitative and qualitative results show that this achieves better performance. To deal with the imbalanced learning problem, several strategies have been tested. The proposed patch-based attention module shows its effectiveness by improving the performance drastically. For future directions, we would like to try more backbones, as well as investigate more attention-based learning methods for point cloud data.

\section*{Acknowledgements}
The project AgiProbot is funded by the Carl Zeiss Foundation.

{\small
\bibliographystyle{ieee_fullname}
\bibliography{egbib}
}

\end{document}